\documentclass[10pt,twocolumn,letterpaper]{article}
\usepackage[pagenumbers]{iccv} 

\usepackage[utf8]{inputenc} 
\usepackage[T1]{fontenc}    
\usepackage{url}            
\usepackage{booktabs}       
\usepackage{amsfonts}       
\usepackage{nicefrac}       
\usepackage{microtype}      

\usepackage{graphicx}
\usepackage{amsmath}
\usepackage{amssymb}

\usepackage{diagbox}
\usepackage{multicol}
\usepackage{enumerate}
\usepackage{times}
\usepackage{epsfig}
\usepackage{threeparttable}
\usepackage{enumitem}
\usepackage{multirow}
\usepackage{color}
\usepackage{array}
\usepackage{setspace}
\usepackage{makecell}
 \usepackage{indentfirst}

\definecolor{iccvblue}{rgb}{0.21,0.49,0.74}
\usepackage[pagebackref,breaklinks,colorlinks,allcolors=iccvblue]{hyperref}

\title{STD-GS: Exploring Frame-Event Interaction for SpatioTemporal-Disentangled \\ Gaussian Splatting to Reconstruct High-Dynamic Scene}

\author{Hanyu Zhou\textsuperscript{\rm 1, 2}\thanks{These authors contributed equally, and this work was finished at Huazhong University of Science and Technology.}, Haonan Wang\textsuperscript{\rm 1}\footnotemark[1], Haoyue Liu\textsuperscript{\rm 1}, Yuxing Duan\textsuperscript{\rm 1}, Luxin Yan\textsuperscript{\rm 1}\thanks{Corresponding authors.}, Gim Hee Lee\textsuperscript{\rm 2}\footnotemark[2]\\
  \textsuperscript{\rm 1} National Key Lab of Multispectral Information Intelligent Processing Technology\\
  School of Artificial Intelligence and Automation, Huazhong University of Science and Technology\\
  \textsuperscript{\rm 2} School of Computing, National University of Singapore\\
  {\tt\small {\{hy.zhou, gimhee.lee\}}@nus.edu.sg} ~~~~~~~~
  {\tt\small yanluxin@hust.edu.cn}
 }

\usepackage[sort&compress]{natbib}
\begin{document}
\maketitle

\begin{abstract}
High-dynamic scene reconstruction aims to represent static background with rigid spatial features and dynamic objects with deformed continuous spatiotemporal features. Typically, existing methods adopt unified representation model (e.g., Gaussian) to directly match the spatiotemporal features of dynamic scene from frame camera. However, this unified paradigm fails in the potential discontinuous temporal features of objects due to frame imaging and the heterogeneous spatial features between background and objects. To address this issue, we disentangle the spatiotemporal features into various latent representations to alleviate the spatiotemporal mismatching between background and objects. In this work, we introduce event camera to compensate for frame camera, and propose a spatiotemporal-disentangled Gaussian splatting framework for high-dynamic scene reconstruction. As for dynamic scene, we figure out that background and objects have appearance discrepancy in frame-based spatial features and motion discrepancy in event-based temporal features, which motivates us to distinguish the spatiotemporal features between background and objects via clustering. As for dynamic object, we discover that Gaussian representations and event data share the consistent spatiotemporal characteristic, which could serve as a prior to guide the spatiotemporal disentanglement of object Gaussians. Within Gaussian splatting framework, the cumulative scene-object disentanglement can improve the spatiotemporal discrimination between background and objects to render the time-continuous dynamic scene. Extensive experiments have been performed to verify the superiority of the proposed method.
\end{abstract}

\begin{figure}
  \setlength{\abovecaptionskip}{5pt}
  \setlength{\belowcaptionskip}{-5pt}
  \centering
   \includegraphics[width=0.99\linewidth]{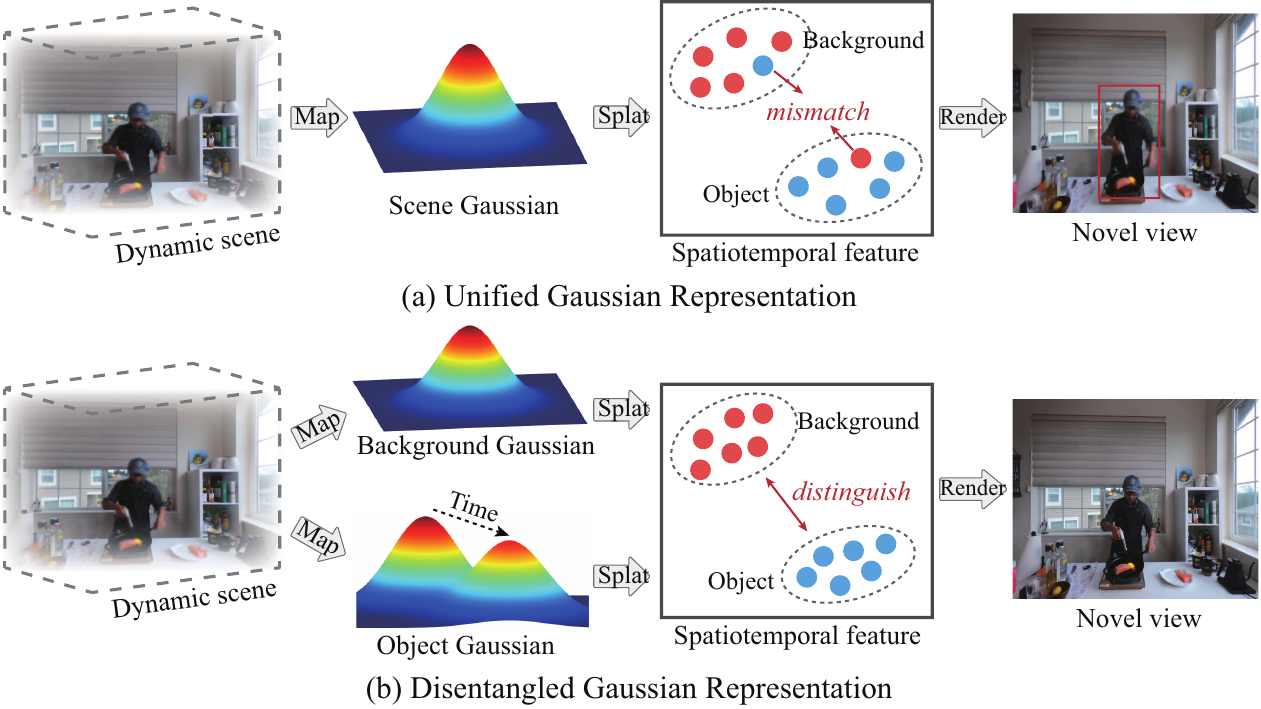}
   \caption{Illustration of Gaussian splatting paradigms for scene reconstruction. Unified Gaussian representation directly model the spatiotemporal features of dynamic scene using the Gaussian with same form, while may suffer the feature mismatch due to heterogeneous feature distribution between static background and dynamic objects. In contrast, we explore a disentangled Gaussian representation to disentangle dynamic scene into various latent Gaussians for distinguishing the spatiotemporal features between background and objects. In this work, we introduce event camera to assist frame camera, and propose a spatiotemporal-disentangled Gaussian splatting framework for high-dynamic scene reconstruction.
   }
   \label{Fig:Paradigm}
\end{figure}

\section{Introduction}
\label{sec:intro}
Traditional scene reconstruction \cite{mildenhall2021nerf, kerbl20233d} aims to represent the 3D static scene with rigid spatial features from 2D frame images. Unlike static scene, dynamic scene includes dynamic objects with deformed continuous spatiotemporal features, which are mixed with the spatial features of the static background, limiting existing scene reconstruction methods (\emph{e.g.}, Gaussian splatting \cite{kerbl20233d}). As the motion patterns of dynamic objects intensify, distinguishing the spatiotemporal features of the objects and the spatial features of the background becomes more challenging in such high-dynamic scenarios. In this work, we aim to propose a novel Gaussian splatting method for high-dynamic scene reconstruction.

Typically, existing dynamic scene reconstruction methods \cite{fang2022fast, yang2024real, lin2024gaussian, wu20244d, wang2024shape} mainly adopt unified representation model (\emph{e.g.}, 3D Gaussian \cite{kerbl20233d}) to directly match the spatiotemporal features of dynamic scene captured from 2D frame images in Fig. \ref{Fig:Paradigm} (a).
For example, Wang \emph{et al.} \cite{wang2024shape} matched multiple key points in 2D images via tracking, and then directly generated multiple 3D Gaussians centered on these points to model dynamic scene. However, these unified representation methods suffer two problems. First, limited by frame imaging, the motion patterns of high-dynamic objects can lead to the potential texture blurring caused by long exposure and large displacement due to low frame rate, thus causing the discontinuity of temporal features. Second, dynamic objects are deformed while static background is rigid, thereby bringing in the heterogeneous nature of spatial features between background and objects. The two difficulties will exacerbate the feature mismatching in high-dynamic scenes. Therefore, \emph{modeling the intrinsic spatiotemporal nature of background and objects is crucial for high-dynamic scene reconstruction}.

To address this issue, we disentangle the spatiotemporal features of high-dynamic scene into various latent Gaussian representations to alleviate the spatiotemporal feature mismatching in Fig. \ref{Fig:Paradigm} (b). Regarding dynamic scene, our motivation is to decompose dynamic scene into static background with rigid spatial features and dynamic objects with deformed continuous spatiotemporal features. Considering the limitation of frame imaging, we introduce event camera \cite{Gallego2019EventBasedVA}, and figure out that background and objects have the appearance discrepancy in the frame-based spatial features and the motion discrepancy in the event-based temporal features. This motivates us to distinguish the spatiotemporal features of background and objects according to the appearance and motion information. Regarding dynamic object, we focus on the spatiotemporal characteristic representation of object Gaussians, where we discover that the spatial color and temporal deformation of Gaussians are consistent with the spatial brightness and temporal flow of event data. This inspires us to take the consistent nature between Gaussians and events as a guide prior to model the spatiotemporal characteristics of dynamic objects. Therefore, the cumulative scene-object disentanglement can improve the spatiotemporal feature discrimination of dynamic scene.

\begin{figure*}
  \setlength{\abovecaptionskip}{5pt}
  \setlength{\belowcaptionskip}{-5pt}
  \centering
   \includegraphics[width=1.0\linewidth]{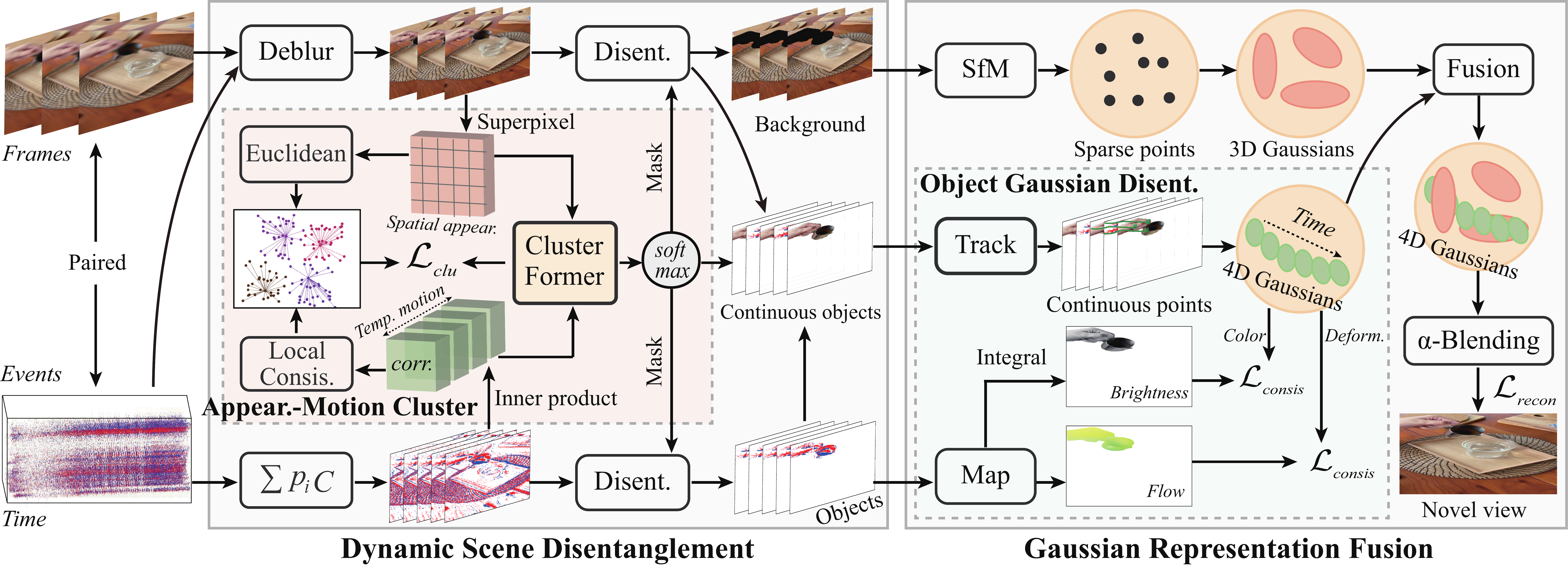}
   \caption{The architecture of the STD-GS mainly contains dynamic scene disentanglement (DSD) and Gaussian representation fusion (GRF). In DSD, we extract the spatial appearance feature from frame images and the temporal motion feature from event stream, and construct a learnable appearance-motion clustering approach to disentangle dynamic scene into background and objects. In GRF, we utilize event data to disentangle the spatiotemporal characteristic of object Gaussians, and fuse background and object Gaussians to render novel-view images.
   }
   \label{Fig:Framework}

\end{figure*}

In this work, we propose a novel \textbf{s}patio\textbf{t}emporal-\textbf{d}isentangled \textbf{G}aussian \textbf{s}platting (STD-GS) framework for high-dynamic scene reconstruction in Fig. \ref{Fig:Framework}, including dynamic scene disentanglement (DSD) and Gaussian representation fusion (GRF).
Specifically, we first obtain the paired frame-event data. In DSD module, we use superpixel \cite{achanta2012slic} to divide the frame image into many local patches as spatial appearance features, and use the inner product to calculate the correlations of events as temporal motion features. Next, we design a learnable appearance-motion clustering network to distinguish the spatiotemporal features of static background and dynamic objects. In GRF module, we first use EKF \cite{einicke1999robust} to update the temporal features of the same dynamic object to generate continuous tracking points, and then represent the static background as 3D Gaussian and the dynamic object as 4D Gaussian centered on these tracking points. We further construct the color-brightness consistency and deformation-flow consistency between the Gaussian and the event to learn the Gaussian spatiotemporal characteristic of the dynamic object. Finally, we use shadow-based weights to fuse the Gaussians of background and objects for rendering. Under this unified framework, the proposed spatiotemporal-disentangled Gaussian splatting can progressively represent the spatiotemporal patterns from scene-level to object-level, thus achieving time-continuous dynamic scene reconstruction. Overall, the main contributions are as follows:
\begin{itemize}[leftmargin=10pt]
\item We introduce event camera to compensate for frame camera, and propose a novel spatiotemporal-disentangled Gaussian splatting framework, which can explicitly represent spatiotemporal patterns to achieve time-continuous high-dynamic reconstruction.

\item We reveal the spatiotemporal discrepancy of dynamic scene and the spatiotemporal consistency of dynamic object, which motivates us to propose a cumulative scene-object disentanglement pipeline to improve the spatiotemporal discrimination of dynamic scene.

\item We build a optical coaxial device to collect the pixel-aligned frame-event dataset, and conduct extensive experiments to verify the superiority of the proposed method.
\end{itemize}

\section{Related Work}
\label{sec:related_work}
\noindent
\textbf{3D Scene Reconstruction.}
The previous 3D scene representations are divided into matching, point, voxel and mesh. Matching-based methods \cite{snavely2006photo, mur2015orb} directly extract key points from adjacent frames, which is projected to sparse point clouds. Point-based methods \cite{aliev2020neural, kopanas2021point, ruckert2022adop} adopt multi-view stereo strategy to estimate the disparity for dense point cloud. However, this point-based representation is limited due to local holes between independent points. Voxel-based methods \cite{guo2019relightables, kerbl20233d, li2017robust} are proposed to represent the scene as multiple fixed-shape cubes to ensure the integrity of the 3D scene structure, while encountering the local feature confusion of different targets. Mesh-based methods \cite{broxton2020immersive, collet2015high, guo2015robust, su2020robustfusion} further transform cubes into variable-shape triangular grids to model the scene, facilitating the fine-grained representation. Recently, Neural Radiance Field (NeRF) methods \cite{mildenhall2021nerf, park2021hypernerf} build on continuous scene representation, optimizing a Multi-Layer Perceptron using volumetric ray-marching for novel view synthesis of captured scenes. 3D Gaussian splatting \cite{kerbl20233d} is a new paradigm of scene representation, which uses a set of 3D Gaussian models to explicitly represent the 3D scene with geometric properties. Despite these 3D scene reconstruction methods have made great progress in static scenes, rendering for dynamic scenes remains challenging. In this work, our purpose is to reconstruct spatiotemporally varying dynamic scene with static background and dynamic objects.

\noindent
\textbf{Dynamic Scene Reconstruction.}
Dynamic scene additionally contain independent dynamic objects with deformed continuous spatiotemporal features, which are mixed with the spatial features of static background. Typically, existing dynamic scene reconstruction methods \cite{fang2022fast, yang2024real, lin2024gaussian, wu20244d, wang2024shape, yang2023emernerf, yang2023emernerf, li2024spacetime, zhu2024motiongs, rudnev2023eventnerf, qi2023e2nerf, ma2023deformable} mainly use unified representation model to directly characterize the spatiotemporal features of dynamic scenes from image sequences. However, these methods have two limitations. First, limited by the long exposure and low frame rate of frame camera, the violent motion patterns of high-dynamic objects can bring in potentially blurry textures and large displacements, thus causing the discontinuity of temporal features. Second, dynamic objects are deformed while static background is rigid, leading to the heterogeneous nature of spatial features between background and objects. The two limitations exacerbate the invalid matching of spatiotemporal features in high-dynamic scenes. Therefore, we focus on modeling the intrinsic spatiotemporal features of dynamic scenes and objects, and disentangle the spatiotemporal features into various latent representation spaces, thereby representing static background with rigid spatial features and dynamic objects with deformed continuous spatiotemporal features for high-dynamic scene reconstruction.

\noindent
\textbf{Gaussian Splatting with Event.}
Gaussian splatting (GS) technology adopts 3D Gaussian model to represent scenes from multiple frame images, which has been applied for novel view synthesis. Frame-based GS methods \cite{lin2024gaussian, li2024spacetime, wu20244d} have made great progress in static scenes and low dynamic scenes, while still limited by the imaging limitation of frame camera in high dynamic scenes, \emph{i.e.}, long exposure and low frame rate. In contrast, the event camera is a neuromorphic sensor, which asynchronously triggers events based on changes in brightness, providing the advantage of high temporal resolution. Recently, several GS methods \cite{xiong2024event3dgs, deguchi2024e2gs} introduce event camera to assist frame camera in better representing the spatiotemporal features of the scene. However, these methods simply use event data to correct the frame data at the original imaging level, lacking a deeper modeling of the complex spatiotemporal features of high-dynamic scenes using multimodal knowledge. To address this, we further distinguish the spatiotemporal discrepancy in dynamic scenes by leveraging the frame-event multimodal dominant knowledge for better background-object Gaussian representation, and propose a novel spatiotemporal-disentangled Gaussian splatting method for high-dynamic scene reconstruction.

\section{Spatiotemporal-Disentangled Gaussian Splat}
\label{sec:method}
\subsection{Overall Framework}
In high-dynamic scenes, limited by frame imaging, there exists the temporal feature discontinuity of objects and the heterogeneous nature of spatial features between background and objects, resulting in that it is difficult to model the spatiotemporal features. To this end, we introduce event camera to assist frame camera, and propose a multimodal spatiotemporal-disentangled Gaussian splatting framework for high-dynamic scene reconstruction in Fig. \ref{Fig:Framework}, including dynamic scene disentanglement and Gaussian representation fusion. The former is to decompose spatiotemporal features of the scene, while the latter is to further decompose spatiotemporal characteristics of the object and then fuse them into the whole scene Gaussians for rendering. In dynamic scene disentanglement, we design a learnable appearance-motion clustering approach to distinguish the spatiotemporal features of the dynamic scene using frame and event as input, thus disentangling the dynamic scene into the static background and the dynamic object. In Gaussian representation fusion, we map the background and object to Gaussian representation, and then generate the continuous tracking points as the center of the object Gaussians, where we utilize the spatiotemporal consistency between Gaussians and events to learn the Gaussian spatiotemporal characteristics of the dynamic object. Next, we apply the shadow-based weights to fuse the Gaussians of the background and object for rendering novel view. Therefore, the cumulative scene-object disentanglement can improve the representation of spatiotemporal patterns, promoting the fusion of spatiotemporal Gaussians for time-continuous dynamic scene reconstruction.

\begin{figure}
  \setlength{\abovecaptionskip}{5pt}
  \setlength{\belowcaptionskip}{-5pt}
  \centering
   \includegraphics[width=0.99\linewidth]{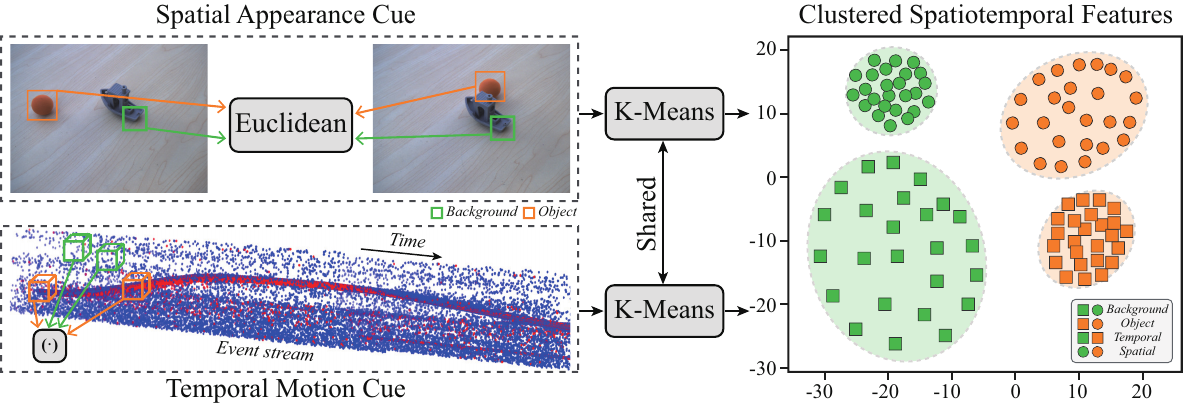}
   \caption{Spatiotemporal discrepancy between the background and object. In imaging level, the background is rigid and discontinuous, while the object is deformed and continuous. In feature level, spatial features of the background are clustered while those of the object are relatively divergent, and temporal features of the background are discrete while those of the object are clustered.
   }
   \label{Fig:Cluster}

\end{figure}

\subsection{Dynamic Scene Disentanglement}
Long exposure and low frame rate of frame camera limit the continuity of temporal features of high-dynamic objects, and the heterogeneous nature of spatial features between background and objects also exacerbates the feature mismatching of dynamic scene. Therefore, we introduce event camera to assist frame camera in modeling temporal information, and disentangle dynamic scene by distinguishing the spatiotemporal features between background and objects.

\noindent
\textbf{Spatiotemporal Discrepancy of Scene.}
As for the high-dynamic motion pattern, frame camera provides dense spatial appearance but discontinuous temporal motion, while event camera provides sparse spatial appearance but continuous temporal motion. This naturally leads us to consider leveraging the dominant knowledge of frame and event data to analyze the spatiotemporal feature discrepancy between background and objects in dynamic scenes. Note that we manually divide the regions of background and objects as candidate regions to match. For the spatial feature, we segment the candidate regions of frame images into superpixels as spatial appearance features, and use Euclidean distance to match the appearance features between different frame images. For the temporal feature, we perform inner product on the candidate regions of event stream to generate the correlation volumes as temporal motion features, and utilize the local consistency (\emph{i.e.}, uniform motion within a short period) to match these motion features. Furthermore, we take K-Means clustering to obtain the spatiotemporal feature distributions of the background and object, and visualize them via t-SNE in Fig. \ref{Fig:Cluster}. In the imaging level, the background is rigid while the object is deformed in frame images, and the background is discontinuous while the object is continuous in event stream. In the feature level, the spatial appearance features of the background are clustered while those of the object are relatively divergent, and the temporal motion features of the background are globally discrete while those of the object are clustered. Therefore, we perform clustering on frame-based appearance features and event-based motion features to achieve dynamic scene disentanglement.

\noindent
\textbf{Learnable Appearance-Motion Clustering.}
For the discrimination between background and objects, appearance clustering depends on frame-based spatial information while motion clustering is determined by event-based temporal information. In order to enable end-to-end spatiotemporal feature discrimination, we propose a learnable appearance-motion clustering approach that combines knowledge and model. Referring to GEM \cite{zhang2023generalizing}, we first utilize event data to restore potential blurry textures of the high-dynamic object in frame images. Regarding prior knowledge, we adopt the same strategy of modeling the spatiotemporal feature discrepancy as shown in Fig. \ref{Fig:Cluster}, where we use superpixel patches from frame images as spatial appearance features and correlation volumes from event stream as temporal motion features. Then, we apply K-Means algorithm to cluster these spatiotemporal features corresponding to static background and dynamic objects, as pseudo-labels to guide subsequent network learning. Regarding deep model, we introduce the ClusterFormer network \cite{liang2024clusterfomer} to achieve recursive updates of cluster centers via clustering consistency loss:
\begin{equation}\small
  \setlength\abovedisplayskip{2pt}
  \setlength\belowdisplayskip{2pt}
\begin{aligned}
\mathcal{L}_{clu} = \sum\nolimits_{i, j} ||x_i - c_{y_i}|| + ||f(x_i) - f(x_j)|| \cdot S(y_i, y_j),
 \label{eq:clustering}
 \end{aligned}
\end{equation}
where $x_i$ is the $i$ sample, and $c_{y_i}$ is the center, $y_i$ is the clustered label from K-Means clustering. $f(x)$ is the feature from ClusterFormer network, $S(\cdot)$ is used to define the relationship between the labels of two samples. This knowledge-and-model-based clustering solution could enhance the representation learning capability of spatiotemporal features in dynamic scenes, making the disentangling process of static background and dynamic objects more interpretable.

\begin{figure}
  \setlength{\abovecaptionskip}{5pt}
  \setlength{\belowcaptionskip}{-5pt}
  \centering
   \includegraphics[width=0.99\linewidth]{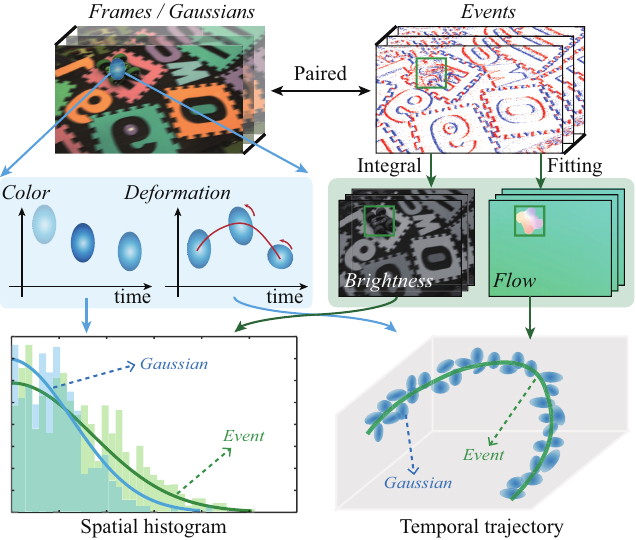}
   \caption{Spatiotemporal consistency between Gaussians and events. As for the same object, the spatial color of Gaussians is similar to the spatial brightness of events, and the temporal deformation of Gaussians aligns with the temporal flow of events.
   }
   \label{Fig:Consistency}

\end{figure}

\subsection{Gaussian Representation Fusion}
Dynamic scene disentanglement could distinguish static background with spatial features and dynamic objects with spatiotemporal features, while the decomposed dynamic objects cannot be directly used to scene reconstruction. The main reason is that the dynamic object consists of frame-based spatial features and event-based temporal features, and the spatiotemporal features with different modalities make it difficult to render the dense and continuous object. Therefore, we further disentangle the dynamic object into the same representation to model the spatiotemporal characteristic, and then fuse the representations of background and objects for rendering dynamic scene.

\noindent
\textbf{Object Gaussian Disentanglement.}
To guarantee the temporal continuity of dynamic objects, we first take the correlation volume of the dynamic object from event data as the template to match, and then we use EKF \cite{einicke1999robust} to track the dynamic object in the event data along the temporal dimension, thereby updating a series of tracking points. Next, we represent the decomposed multi-view static background into 3D Gaussians $\mathcal{G}_s$ with color $\mathcal{C}_s$ and continuous dynamic objects with temporal scale into 4D Gaussians $\mathcal{G}_d$ with color $\mathcal{C}_d$ centered on these updated tracking points.
Considering that Gaussian representation reconstructs the scene without changing the intrinsic nature, we argue that the event data and Gaussian representation of the dynamic object are different in representation form, but their spatiotemporal characteristics are expected to be consistent. To validate this insight, we further disentangle the spatiotemporal characteristics of the Gaussian (\emph{i.e.}, spatial color and temporal deformation), which are compared with the characteristics of the events (\emph{i.e.}, spatial brightness and temporal motion) in Fig. \ref{Fig:Consistency}. We can observe that for the same dynamic object, the spatial histogram of Gaussian color is consistent with that of the integral brightness of the events, and the temporal trajectory of Gaussian deformation aligns with that of the event-based optical flow. Motivated by this, we treat the spatiotemporal characteristics of event data as pseudo-labels, and construct a spatiotemporal consistency loss between event data and Gaussian feature representation to constrain the learning of the spatiotemporal characteristics of dynamic objects:
\begin{equation}\small
  \setlength\abovedisplayskip{2pt}
  \setlength\belowdisplayskip{2pt}
\begin{aligned}
\mathcal{L}_{consis} = \sum\nolimits||\mathcal{C}_d - L_{ev}||_1 + ||\mathcal{F}(\mathcal{G}_d)-F_{ev}||_1,
 \label{eq:consistency}
 \end{aligned}
\end{equation}
where $\mathcal{F}$ is the Gaussian deformation operator. $L_{ev}$ and $F_{ev}$ denote spatial brightness and temporal flow from event data, respectively. Therefore, the cumulative disentanglement from dynamic scene to dynamic object can enhance the discrimination of spatiotemporal features in dynamic scenes.

\begin{table*}\footnotesize
    \setlength{\abovecaptionskip}{5pt}
    \setlength\tabcolsep{1.5pt}
    \setlength{\belowcaptionskip}{-5pt}

  \centering
  \renewcommand\arraystretch{1.1}
  \begin{tabular}{ccccccccccc}
  \Xhline{1pt}
  \multicolumn{2}{c|}{\multirow{2}{*}{Method}}&
  \multicolumn{5}{c|}{\multirow{1}{*}{Static scene reconstruction method}}&
  \multicolumn{4}{c}{\multirow{1}{*}{Dynamic scene reconstruction method}} \\

  \cline{3-11}
  \multicolumn{2}{c|}{}& \multicolumn{1}{c|}{\multirow{1}{*}{3DGS \cite{kerbl20233d}}} & \multicolumn{1}{c|}{\multirow{1}{*}{3DGS w/ De}} & \multicolumn{1}{c|}{\multirow{1}{*}{EventNeRF \cite{rudnev2023eventnerf}}} & \multicolumn{1}{c|}{\multirow{1}{*}{E2NeRF \cite{qi2023e2nerf}}} & \multicolumn{1}{c|}{\multirow{1}{*}{E2GS \cite{deguchi2024e2gs}}} &
  \multicolumn{1}{c|}{\multirow{1}{*}{TiNeuVox \cite{fang2022fast}}}&
  \multicolumn{1}{c|}{\multirow{1}{*}{TiNeuVox w/ De}}&
  \multicolumn{1}{c|}{\multirow{1}{*}{4DGS \cite{wu20244d}}}&
  \multicolumn{1}{c}{\multirow{1}{*}{\textbf{Ours}}} \\

  \hline
  \multicolumn{2}{c|}{\multirow{1}{*}{Input}}&  \multicolumn{1}{c|}{Frame} & \multicolumn{1}{c|}{Frame} & \multicolumn{1}{c|}{Event}& \multicolumn{1}{c|}{Frame+Event}&  \multicolumn{1}{c|}{Frame+Event}&  \multicolumn{1}{c|}{Frame} & \multicolumn{1}{c|}{Frame} & \multicolumn{1}{c|}{Frame} & \multicolumn{1}{c}{\textbf{Frame+Event}}\\

  \hline
   \multicolumn{1}{c|}{\multirow{3}{*}{\makecell{Event-\\HyperNeRF}}}& \multicolumn{1}{c|}{PSNR $\uparrow$} & \multicolumn{1}{c|}{17.08} & \multicolumn{1}{c|}{20.05} & \multicolumn{1}{c|}{15.60}& \multicolumn{1}{c|}{19.42}&  \multicolumn{1}{c|}{20.01}&  \multicolumn{1}{c|}{18.11} & \multicolumn{1}{c|}{20.34} & \multicolumn{1}{c|}{18.29} & \multicolumn{1}{c}{\textbf{23.57}}\\

  \multicolumn{1}{c|}{} & \multicolumn{1}{c|}{SSIM $\uparrow$} & \multicolumn{1}{c|}{0.54} & \multicolumn{1}{c|}{0.63} &
  \multicolumn{1}{c|}{0.48}& \multicolumn{1}{c|}{0.60}&
  \multicolumn{1}{c|}{0.62}&  \multicolumn{1}{c|}{0.54} &
  \multicolumn{1}{c|}{0.61} & \multicolumn{1}{c|}{0.55} & \multicolumn{1}{c}{\textbf{0.69}}\\

  \multicolumn{1}{c|}{} & \multicolumn{1}{c|}{LPIPS $\downarrow$} & \multicolumn{1}{c|}{0.51} & \multicolumn{1}{c|}{0.48} &
  \multicolumn{1}{c|}{0.53}& \multicolumn{1}{c|}{0.48}&
  \multicolumn{1}{c|}{0.45}&  \multicolumn{1}{c|}{0.49} &
  \multicolumn{1}{c|}{0.45} & \multicolumn{1}{c|}{0.47} & \multicolumn{1}{c}{\textbf{0.38}}\\

  \hline
   \multicolumn{1}{c|}{\multirow{3}{*}{HD-CED}}& \multicolumn{1}{c|}{PSNR $\uparrow$} & \multicolumn{1}{c|}{16.83} & \multicolumn{1}{c|}{17.75} & \multicolumn{1}{c|}{15.94}& \multicolumn{1}{c|}{18.85}&  \multicolumn{1}{c|}{18.73}&  \multicolumn{1}{c|}{19.45} & \multicolumn{1}{c|}{20.56} & \multicolumn{1}{c|}{20.60} & \multicolumn{1}{c}{\textbf{27.12}}\\

  \multicolumn{1}{c|}{} & \multicolumn{1}{c|}{SSIM $\uparrow$} & \multicolumn{1}{c|}{0.56} & \multicolumn{1}{c|}{0.58} &
  \multicolumn{1}{c|}{0.49}& \multicolumn{1}{c|}{0.61}&
  \multicolumn{1}{c|}{0.62}&  \multicolumn{1}{c|}{0.64} &
  \multicolumn{1}{c|}{0.67} & \multicolumn{1}{c|}{0.67} & \multicolumn{1}{c}{\textbf{0.88}}\\

  \multicolumn{1}{c|}{} & \multicolumn{1}{c|}{LPIPS $\downarrow$} & \multicolumn{1}{c|}{0.49} & \multicolumn{1}{c|}{0.40} &
  \multicolumn{1}{c|}{0.53}& \multicolumn{1}{c|}{0.39}&
  \multicolumn{1}{c|}{0.38}&  \multicolumn{1}{c|}{0.36} &
  \multicolumn{1}{c|}{0.32} & \multicolumn{1}{c|}{0.31} & \multicolumn{1}{c}{\textbf{0.25}}\\

  \Xhline{1pt}
  \end{tabular}
  \caption{Quantitative results on Event-HyperNeRF and HD-CED datasets. ``w/ De'' is deblurring image before scene reconstruction.}
   \label{Tab:Public_Comparison}
\end{table*}

\begin{figure*}
  \setlength{\abovecaptionskip}{5pt}
  \setlength{\belowcaptionskip}{-5pt}
  \centering
   \includegraphics[width=0.99\linewidth]{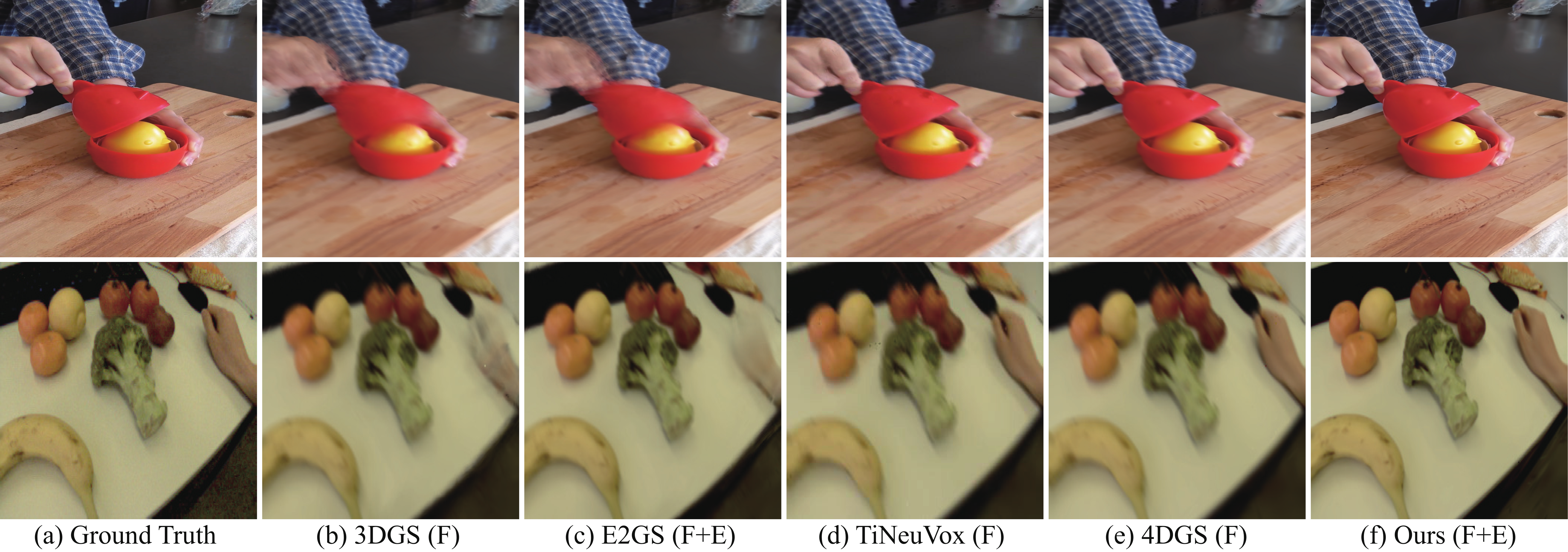}
   \caption{Visual comparison of novel view synthesis on Event-HyperNeRF and HD-CED datasets. ``F'' denotes frame and ``E'' denotes event.
   }
   \label{Fig:Public_Comparison}

\end{figure*}

\noindent
\textbf{Gaussian Fusion for Splatting.}
The decoupled spatiotemporal features need to be further fused for novel view synthesis. Note that the representation fusion of static background and dynamic objects includes both Gaussian distribution fusion and color fusion. Given a view matrix $M=[R, T]$ and a timestamp $t$, considering the presence of shadow between background and objects, we introduce a MLP to learn the overlap probability $\rho$ between them, as a weight to fuse background Gaussians $\mathcal{G}_s$ and object Gaussians $\mathcal{G}_d$ to obtain a new Gaussian distribution in the novel view and timestamp:
\begin{equation}\small
  \setlength\abovedisplayskip{2pt}
  \setlength\belowdisplayskip{2pt}
\begin{aligned}
 \mathcal{\hat{G}} = (1 - \rho) \cdot \mathcal{H}(M, \mathcal{G}_s) + \rho \cdot \mathcal{H}(M,  \mathcal{G}_d(t)),
 \label{eq:gaussian_fusion}
 \end{aligned}
\end{equation}
where $\mathcal{H}$ denotes the differential splatting \cite{yifan2019differentiable}, and we also apply the same strategy for the color fusion as follows:
\begin{equation}\small
  \setlength\abovedisplayskip{2pt}
  \setlength\belowdisplayskip{2pt}
\begin{aligned}
   \mathcal{\hat{C}} = \frac{\sigma_s}{\sigma_s+\sigma_d} \cdot (1 - \rho) \cdot \mathcal{C}_s + \frac{\sigma_d}{\sigma_s+\sigma_d} \cdot \rho \cdot \mathcal{C}_d,
 \label{eq:color_fusion}
 \end{aligned}
\end{equation}
where $\sigma_s$, $\sigma_d$ denote the density of background and object Gaussians, respectively. Finally, we utilize $\alpha$-blending \cite{lassner2021pulsar, kerbl20233d} to render the fused Gaussian $\mathcal{\hat{G}}$ and fused color $\mathcal{\hat{C}}$ to obtain a novel-view image $\hat{I}$ with reconstruction loss \cite{kerbl20233d}:
\begin{equation}\small
  \setlength\abovedisplayskip{2pt}
  \setlength\belowdisplayskip{2pt}
\begin{aligned}
\mathcal{L}_{recon} = ||\hat{I} - I_{gt}||_1 + \mathcal{L}_{ssim},
 \label{eq:reconstruction}
 \end{aligned}
\end{equation}
where $\mathcal{L}_{ssim}$ measures the structural similarity between reconstructed image and GT image. In this way, the proposed spatiotemporal-disentangled Gaussian splatting can reconstruct a time-continuous high-dynamic scene.

\subsection{Optimization and Implementation Details}
Consequently, the total objective for the proposed Gaussian splatting framework is written as follows:
\begin{equation}\small
  \setlength\abovedisplayskip{2pt}
  \setlength\belowdisplayskip{2pt}
\begin{aligned}
\mathcal{L} = \lambda_{1} \mathcal{L}_{recon}
              + \lambda_{2} \mathcal{L}_{clu}
              + \lambda_{3} \mathcal{L}_{consis},
 \label{eq:total_loss}
 \end{aligned}
\end{equation}
where ${[\lambda_{1}, \lambda_{2}, \lambda_{3}]}$ are the weights that control the importance of the related losses. The first term is to ensure the novel-view image synthesis capability of the Gaussian splatting framework. The second term disentangles dynamic scene by learning the spatiotemporal feature discrepancy between static background and dynamic objects. The third term further disentangles the dynamic objects by constraining the spatiotemporal characteristic consistency between the Gaussian representation and event data. Under this unified framework, these three losses support the decomposition-fusion learning paradigm, which helps improve the spatiotemporal representation for high-dynamic scenes.

Regarding implementation details, our implementation is primarily based on the PyTorch \cite{paszke2019pytorch} and tested in a single RTX 3090 GPU. During the training phase, we have two steps. First, we use $\mathcal{L}_{recon}$ to train the whole Gaussian splatting framework for initializing scene reconstruction capabilities, including static background and dynamic objects. Second, we use $\mathcal{L}_{clu}$ to constrain the appearance-motion clustering process for dynamic scene disentanglement, and simultaneously use $\mathcal{L}_{consis}$ to fine-tune the Gaussian representation model for dynamic objects to improve their spatiotemporal characteristic representation. During the testing phase, the final inference model only includes ClusterFormer network and Gaussian representation model, thus achieving efficient high-dynamic scene reconstruction.

\begin{table}\scriptsize
    \setlength{\abovecaptionskip}{5pt}
    \setlength\tabcolsep{0.8pt}
    \setlength{\belowcaptionskip}{-5pt}

  \centering
  \renewcommand\arraystretch{1.1}
  \begin{tabular}{ccccccc}
  \Xhline{1pt}
  \multicolumn{2}{c|}{\multirow{1}{*}{Method}}&
  \multicolumn{1}{c|}{\multirow{1}{*}{3DGS \cite{kerbl20233d}}} &
   \multicolumn{1}{c|}{\multirow{1}{*}{E2GS \cite{deguchi2024e2gs}}} & \multicolumn{1}{c|}{\multirow{1}{*}{TiNeuVox \cite{fang2022fast}}} & \multicolumn{1}{c|}{\multirow{1}{*}{4DGS \cite{wu20244d}}} & \multicolumn{1}{c}{\multirow{1}{*}{\textbf{Ours}}} \\

  \hline
  \multicolumn{2}{c|}{\multirow{1}{*}{Input}}&  \multicolumn{1}{c|}{Frame}& \multicolumn{1}{c|}{Frame+Event}& \multicolumn{1}{c|}{Frame}&  \multicolumn{1}{c|}{Frame}& \multicolumn{1}{c}{\textbf{Frame+Event}}\\

  \hline
  \multicolumn{1}{c|}{\multirow{3}{*}{\makecell{Low-\\dynamic}}}& \multicolumn{1}{c|}{PSNR $\uparrow$} & \multicolumn{1}{c|}{21.54}& \multicolumn{1}{c|}{22.83}&  \multicolumn{1}{c|}{26.54}& \multicolumn{1}{c|}{29.36}& \multicolumn{1}{c}{\textbf{31.42}}\\

  \multicolumn{1}{c|}{} & \multicolumn{1}{c|}{SSIM $\uparrow$}  & \multicolumn{1}{c|}{0.70}& \multicolumn{1}{c|}{0.72}&  \multicolumn{1}{c|}{0.77}& \multicolumn{1}{c|}{0.83}& \multicolumn{1}{c}{\textbf{0.89}} \\

  \multicolumn{1}{c|}{} & \multicolumn{1}{c|}{LPIPS $\downarrow$}  & \multicolumn{1}{c|}{0.46}& \multicolumn{1}{c|}{0.42}&  \multicolumn{1}{c|}{0.39}& \multicolumn{1}{c|}{0.36}& \multicolumn{1}{c}{\textbf{0.24}} \\

  \hline
  \multicolumn{1}{c|}{\multirow{3}{*}{\makecell{High-\\dynamic}}}& \multicolumn{1}{c|}{PSNR $\uparrow$} & \multicolumn{1}{c|}{19.86}& \multicolumn{1}{c|}{21.98}&  \multicolumn{1}{c|}{23.16}& \multicolumn{1}{c|}{25.23}& \multicolumn{1}{c}{\textbf{27.68}}\\

  \multicolumn{1}{c|}{} & \multicolumn{1}{c|}{SSIM $\uparrow$}  & \multicolumn{1}{c|}{0.62}& \multicolumn{1}{c|}{0.69}&  \multicolumn{1}{c|}{0.72}& \multicolumn{1}{c|}{0.78}& \multicolumn{1}{c}{\textbf{0.87}} \\

  \multicolumn{1}{c|}{} & \multicolumn{1}{c|}{LPIPS $\downarrow$}  & \multicolumn{1}{c|}{0.47}& \multicolumn{1}{c|}{0.43}&  \multicolumn{1}{c|}{0.43}& \multicolumn{1}{c|}{0.41}& \multicolumn{1}{c}{\textbf{0.26}} \\

  \Xhline{1pt}
  \end{tabular}
  \caption{Quantitative results on the proposed CoFED dataset.}
   \label{Tab:Our_Comparison}
\end{table}

\begin{figure*}
  \setlength{\abovecaptionskip}{5pt}
  \setlength{\belowcaptionskip}{-10pt}
  \centering
   \includegraphics[width=0.99\linewidth]{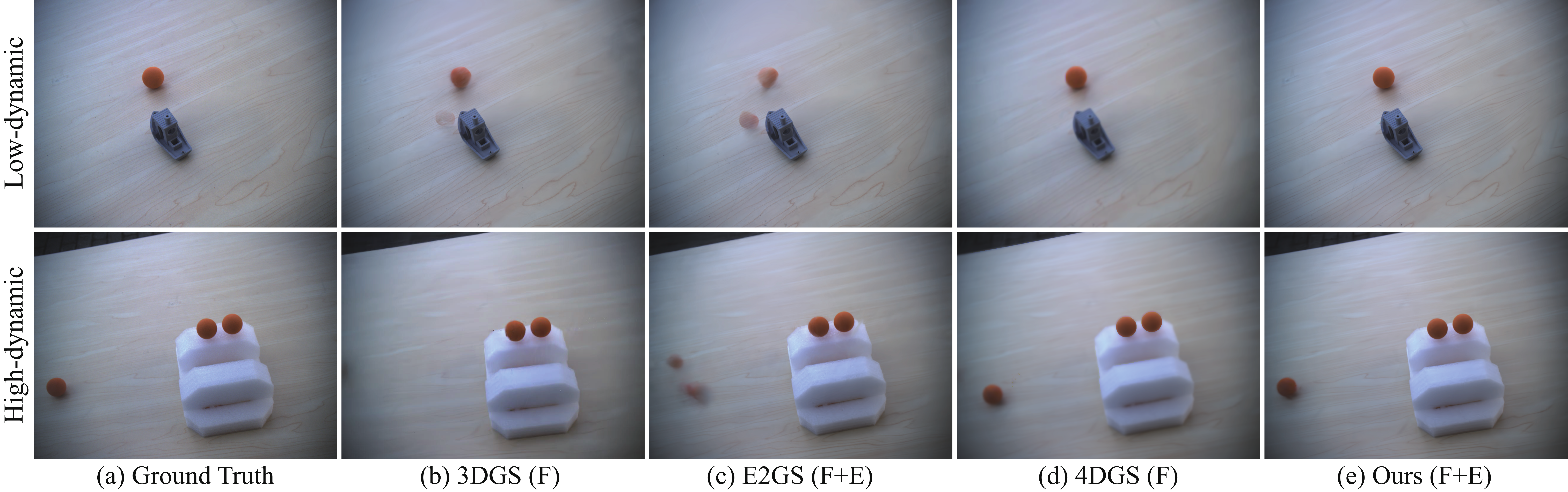}
   \caption{Visual comparison of novel view synthesis on the proposed CoFED dataset. ``F'' denotes frame and ``E'' denotes event.
   }
   \label{Fig:Our_Comparison}

\end{figure*}

\section{Experiments}
\label{sec:experiment}
\subsection{Experiment Setup}
\noindent
\textbf{Dataset.}
We conduct experiments on synthetic (\emph{e.g.}, Event-HyperNeRF) and real (\emph{e.g.}, HD-CED and CoFED) datasets.

\noindent
$\bullet$ \textbf{Event-HyperNeRF.}
HyperNeRF \cite{park2021hypernerf} is a public frame-based dynamic scene reconstruction dataset, where dynamic objects move relatively slowly. To simulate high-dynamic scenes, we perform frame extraction and blur synthesis to introduce spatiotemporal degradation, and v2e model \cite{hu2021v2e} to generate corresponding event data, thus obtaining a synthetic high-dynamic scene reconstruction dataset.

\noindent
$\bullet$ \textbf{HD-CED.}
CED \cite{scheerlinck2019ced} is a real event dataset with a spatial resolution of 346$\times$260. Before this dataset can be used for high-dynamic scene reconstruction comparison, we also perform frame extraction and blur synthesis to simulate high-dynamic scenes, namely HD-CED.

\noindent
$\bullet$ \textbf{CoFED.}
Due to the lack of the real event dataset with high spatial resolution for high-dynamic scene reconstruction, we propose a coaxial frame-event dataset (CoFED, seeing supplementary for details) with a resolution of 1200$\times$624. We build a optical coaxial device with gyroscope to collect the pixel-aligned frame and event data, which covers various motion patterns of high-dynamic objects.

\noindent
\textbf{Comparison Methods.}
We divide comparison methods into unimodal and multimodal categories for static and dynamic scene reconstruction. For static scene, unimodal methods include frame-only (\emph{e.g.}, 3DGS \cite{kerbl20233d}) and event-only (\emph{e.g.}, EventNeRF \cite{rudnev2023eventnerf}) approaches, while multimodal methods (\emph{e.g.}, E2NeRF \cite{qi2023e2nerf} and E2GS \cite{deguchi2024e2gs}) require frame and event data for fusion. For dynamic scene, since the competing methods are mainly based on frame camera, we choose TiNeuVox \cite{fang2022fast} and 4DGS \cite{wu20244d} for comparison. As for the comparison experiments, we have two training strategies for competing methods, one is to directly train the competing methods on the degraded frames; the other is to first performs image restoration (\emph{e.g.}, DeblurGAN \cite{kupyn2018deblurgan}), and then train the methods on the restored results (named as “w/ Deblur”).

\noindent
\textbf{Evaluation Metric.}
We choose peak signal-to-noise ratio (PSNR \cite{hore2010image}), structural similarity index measure (SSIM \cite{wang2004image}), and learned perceptual image patch similarity (LPIPS \cite{zhang2018unreasonable}) as the quantitative evaluation metrics.

\begin{table}\small
    \setlength{\abovecaptionskip}{5pt}
    \setlength\tabcolsep{2pt}
    \setlength{\belowcaptionskip}{-5pt}
  \centering
  \renewcommand\arraystretch{1.1}
  \begin{tabular}{cccc|ccc}
    \Xhline{1pt}
      \multicolumn{2}{c|}{\multirow{1}{*}{Disentangle}} &
      \multicolumn{2}{c|}{\multirow{1}{*}{Fusion}} & \multicolumn{1}{c}{\multirow{2}{*}{PSNR $\uparrow$}}&
      \multicolumn{1}{c}{\multirow{2}{*}{SSIM $\uparrow$}} &
      \multicolumn{1}{c}{\multirow{2}{*}{LPIPS $\downarrow$}}\\
      \cline{1-4}
      \multicolumn{1}{c}{Scene} & \multicolumn{1}{c|}{Object} & \multicolumn{1}{c}{Gaussian} & \multicolumn{1}{c|}{Color} & & & \\
      \hline
      \multicolumn{1}{c}{$\times$} & \multicolumn{1}{c|}{$\times$} & \multicolumn{1}{c}{$\times$} & \multicolumn{1}{c|}{$\times$} & 24.83& 0.75 & 0.42 \\

      \multicolumn{1}{c}{$\surd$} & \multicolumn{1}{c|}{$\times$} & \multicolumn{1}{c}{$\times$} & \multicolumn{1}{c|}{$\times$} & 26.15 & 0.80 & 0.34 \\

      \multicolumn{1}{c}{$\surd$} & \multicolumn{1}{c|}{$\surd$} & \multicolumn{1}{c}{$\times$} & \multicolumn{1}{c|}{$\times$} & 26.89 & 0.83 & 0.30 \\

      \multicolumn{1}{c}{$\surd$} & \multicolumn{1}{c|}{$\surd$} & \multicolumn{1}{c}{$\surd$} & \multicolumn{1}{c|}{$\times$} & 27.26 & 0.85 & 0.28 \\

      \multicolumn{1}{c}{$\surd$} & \multicolumn{1}{c|}{$\surd$} & \multicolumn{1}{c}{$\surd$} & \multicolumn{1}{c|}{$\surd$} & \textbf{27.68} & \textbf{0.87} & \textbf{0.26} \\

       \Xhline{1pt}
  \end{tabular}
  \caption{Effectiveness of spatiotemporal-disentangled framework.}
   \label{Tab:Ablation_Framework}
\end{table}

\subsection{Comparison Experiment}
\noindent
\textbf{Comparison on Synthetic Dataset.}
In Table \ref{Tab:Public_Comparison} and Fig. \ref{Fig:Public_Comparison}, we compare the competing methods with the two training strategies. First, deblurring preprocessing does indeed enhance the performance of frame-based scene reconstruction methods, but it suffers from an upper limitation. Second, multimodal methods outperform unimodal methods, indicating that the complementary knowledge between frame and event modalities helps improve the spatiotemporal feature representation of dynamic scene.

\noindent
\textbf{Comparison on Real Scenes.}
In Table \ref{Tab:Public_Comparison} and Fig. \ref{Fig:Public_Comparison}, we compare the competing methods in more complex dynamic scenes with real event data. First, event-only methods cannot perform better than frame-only methods. The main reason is that single event data lacks a brightness reference, leading to potential inaccurate absolute brightness of rendered images. Second, multimodal methods outperform unimodal frame-only methods in reconstructing scenes. These two conclusions indicate that event data need to be fused together with frame images for scene reconstruction. Furthermore, the proposed method with spatiotemporal disentanglement performs significantly better than other multimodal methods.

\begin{table}\small
    \setlength{\abovecaptionskip}{5pt}
    \setlength\tabcolsep{3pt}
    \setlength{\belowcaptionskip}{-9pt}
  \centering
  \renewcommand\arraystretch{1.1}
  \begin{tabular}{ccc|ccc}
    \Xhline{1pt}
      $\mathcal{L}_{recon}$ & $\mathcal{L}_{clu}$ & $\mathcal{L}_{consis}$& PSNR $\uparrow$ & SSIM $\uparrow$ & LPIPS $\downarrow$\\
			\hline
  		  $\surd$& $\times$&$\times$& 25.78 & 0.79 & 0.38 \\

  		$\surd$& $\surd$&$\times$& 26.95 & 0.84 & 0.30 \\
  		$\surd$& $\times$&$\surd$& 26.03 & 0.80 & 0.36 \\
  		$\surd$& $\surd$&$\surd$& \textbf{27.68} & \textbf{0.87} & \textbf{0.26} \\
       \Xhline{1pt}
  \end{tabular}
  \caption{Ablation study on main losses.}
   \label{Tab:Loss}
\end{table}

\noindent
\textbf{Comparison on Various Dynamic Scenes.}
In Table \ref{Tab:Our_Comparison} and Fig. \ref{Fig:Our_Comparison}, we compare the performance of various methods for dynamic scene reconstruction under different motion patterns (\emph{e.g.}, low-dynamic and high-dynamic) on the proposed dataset. First, dynamic scene reconstruction methods are superior to static scene reconstruction methods. Second, in high-dynamic scenes, the image rendered by competing methods appears blurry texture and position shift of the dynamic object, while the proposed method performs better.

\subsection{Ablation Study}
\noindent
\textbf{How Spatiotemporal-Disentangled Framework Works.}
In Table \ref{Tab:Ablation_Framework}, we demonstrate the effectiveness of the proposed spatiotemporal-disentangled Gaussian splatting framework, including spatiotemporal disentanglement (\emph{e.g.}, scene and object) and representation fusion (\emph{e.g.}, Gaussian and color). With only spatiotemporal disentanglement, scene reconstruction performance significantly improves. When representation fusion is introduced, scene reconstruction performance is further improved. Therefore, within the proposed framework, spatiotemporal disentanglement plays a primary role in enhancing the feature discrimination of the scene, while representation fusion further refines scene reconstruction.

\noindent
\textbf{Effectiveness of Losses.}
In Table \ref{Tab:Loss}, we conduct ablation studies on the main losses. The reconstruction loss $\mathcal{L}_{recon}$ plays a key role in novel-view image rendering, while the appearance-motion clustering loss $\mathcal{L}_{clu}$ and spatiotemporal consistency loss $\mathcal{L}_{consis}$ improve the dynamic scene reconstruction performance to a certain extent. The three losses support the decomposition-fusion learning paradigm of the proposed method for high-dynamic scene reconstruction.

\begin{table}\small
    \setlength{\abovecaptionskip}{5pt}
    \setlength\tabcolsep{4pt}
    \setlength{\belowcaptionskip}{-5pt}
  \centering
  \renewcommand\arraystretch{1.1}
  \begin{tabular}{c|ccc}
    \Xhline{1pt}
      Training data & PSNR $\uparrow$ & SSIM $\uparrow$ & LPIPS $\downarrow$ \\
      \hline
      w/ Frame & 25.91 & 0.81 & 0.37\\
       w/ Event & 24.15 & 0.73 & 0.43\\
       w/ Frame-Event & \textbf{27.68} & \textbf{0.87} & \textbf{0.26}\\
       \Xhline{1pt}
  \end{tabular}
  \caption{Effect of training data on scene reconstruction.}
   \label{Tab:Ablation_TrainData}
\end{table}

\begin{figure}
  \setlength{\abovecaptionskip}{5pt}
  \setlength{\belowcaptionskip}{-5pt}
  \centering
   \includegraphics[width=0.99\linewidth]{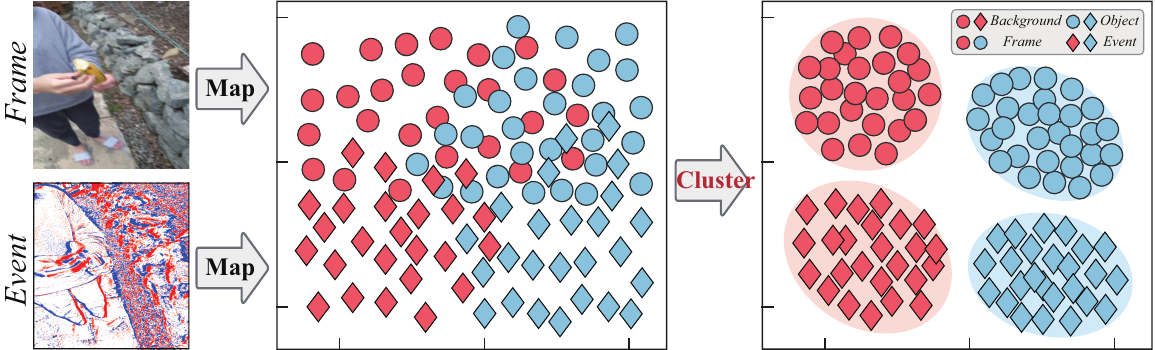}
   \caption{t-SNE visualization of spatiotemporal features. Frame-based spatial features and event-based temporal features are mixed together. After clustering, the spatiotemporal features of background and objects form their own feature set.
   }
   \label{Fig:Discussion_Clustering}

\end{figure}

\noindent
\textbf{Influence of Training data.}
In Table \ref{Tab:Ablation_TrainData}, we discuss the impact of training data (\emph{e.g.}, frame, event and frame-event) on high-dynamic scene reconstruction. First, the effect of event as training data is not good as that of frame on dynamic scene reconstruction. This is because event data just senses the relative varying value of brightness, making it difficult to reflect color information of real world. With frame-event data, the proposed method leverages the complementary knowledge from both modalities to model spatial features from frame images and temporal features from event data, enhancing the performance of high-dynamic scene reconstruction.

\subsection{Discussion}
\noindent
\textbf{Role of Appearance-Motion Clustering.}
We analyze the role of appearance-motion clustering in dynamic scene reconstruction via feature visualization in Fig. \ref{Fig:Discussion_Clustering}. Before clustering, the frame-based spatial features and event-based temporal features for background and objects are scattered and mixed together. After introducing appearance-motion clustering, the spatiotemporal features of background and objects are pushed away. This demonstrates that appearance-motion clustering effectively disentangles the spatiotemporal features of static background and dynamic objects.

\noindent
\textbf{Impact of Event Characteristic for Object Gaussian.}
We discuss the impact of event spatiotemporal characteristics (\emph{i.e.}, spatial brightness and temporal flow) on the object Gaussians in Fig. \ref{Fig:Discussion_Gaussian}. Without event-based spatiotemporal constraint, the distributions of the object Gaussians show spatiotemporal abrupt changes, resulting in color shift and object shift in the rendered image. With event-based spatial brightness, the quality of the rendered image improves significantly, but temporal abrupt changes still exist in the Gaussian distribution. When further introducing event-based temporal flow constraint, the rendered image shows clear structure, and the Gaussian distributions become better.

\noindent
\textbf{Frame Interpolation \emph{v.s.} Temporal Tracking.}
Temporal tracking in the proposed method is designed to ensure the motion continuity of dynamic objects. However, Could this be replaced with frame interpolation instead? To verify this, we compare the effects of frame interpolation (\emph{e.g.}, TimeLens \cite{tulyakov2021time}) and temporal tracking on dynamic scene reconstruction in Table \ref{Tab:Discussion_Continuity}. We can observe that frame interpolation does indeed improve reconstruction performance to some extent, but its metrics are lower than those of temporal tracking. The main reason is that, while frame interpolation can produce continuous visual results, it does not directly model the spatial position and temporal correspondence of dynamic objects. In contrast, temporal tracking better captures the spatiotemporal properties of dynamic objects.

\begin{table}\small
    \setlength{\abovecaptionskip}{5pt}
    \setlength\tabcolsep{3pt}
    \setlength{\belowcaptionskip}{-5pt}
  \centering
  \renewcommand\arraystretch{1.1}
  \begin{tabular}{c|ccc}
    \Xhline{1pt}
      Strategy for continuity & PSNR $\uparrow$ & SSIM $\uparrow$ & LPIPS $\downarrow$ \\
      \hline
      Ours w/o any strategy & 26.25 & 0.81 & 0.33 \\
       Ours w/ frame interpolation & 26.48 & 0.81 & 0.31\\
       Ours w/ temporal tracking & \textbf{27.68} & \textbf{0.87} & \textbf{0.26}\\
       \Xhline{1pt}
  \end{tabular}
  \caption{Discussion on strategies for modeling temporal continuity.}
   \label{Tab:Discussion_Continuity}
\end{table}

\begin{figure}
  \setlength{\abovecaptionskip}{5pt}
  \setlength{\belowcaptionskip}{-5pt}
  \centering
   \includegraphics[width=0.99\linewidth]{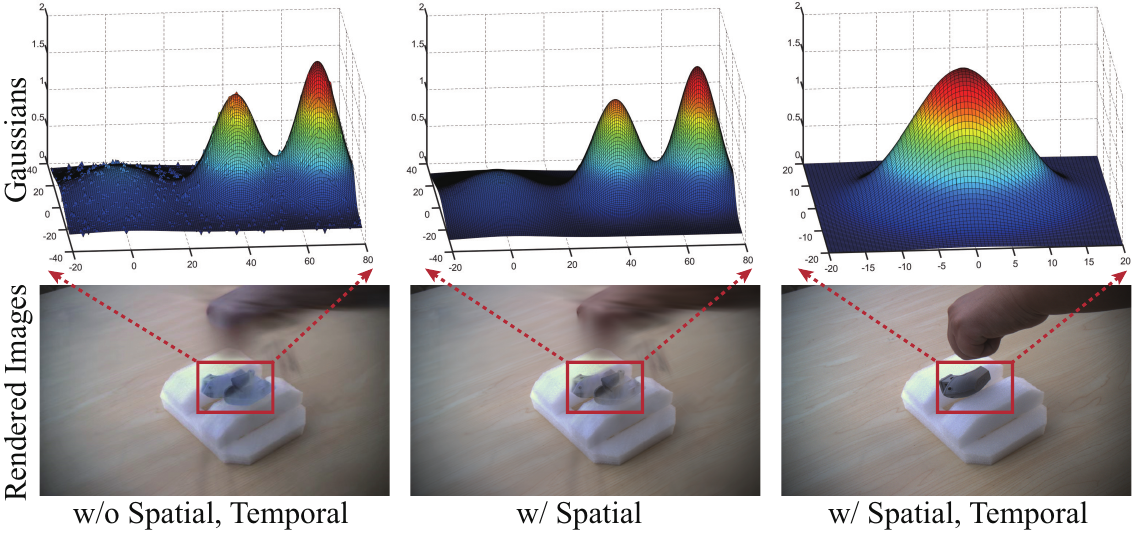}
   \caption{Impact of spatiotemporal characteristics of events. ``Spatial'' is spatial brightness, and ``temporal'' is temporal flow. Spatial brightness of events can smooth Gaussian and restore color shift, while temporal flow of events can alleviate position shift of objects.
   }
   \label{Fig:Discussion_Gaussian}

\end{figure}

\noindent
\textbf{Limitation.}
The proposed method may suffer challenges for the radial dynamic object, which moves to the camera along the z-axis. This is because frame and event cameras can only capture x, y-axis motion patterns, while the radial dynamic object makes the two cameras ineffective for this z-axis special motion pattern, thus causing the spatiotemporal features of the object to resemble those of the background. In the future, we will introduce LiDAR to assist the frame camera in modeling the 3D motion field to distinguish the x, y, z-axis spatiotemporal features of dynamic scene.

\section{Conclusion}
\label{sec:conclusion}
In this work, we introduce event camera to make up for frame camera, and propose a novel multimodal spatiotemporal-disentangled Gaussian splatting framework, which can explicitly represent spatiotemporal patterns via decomposition-fusion learning paradigm to achieve time-continuous high-dynamic scene reconstruction. We design a cumulative scene-object disentanglement pipeline to improve the spatiotemporal discrimination of dynamic scene for rendering. Moreover, we build a optical coaxial device to collect the pixel-aligned frame-event dataset, and demonstrate the superiority of the proposed method. We believe that our work could facilitate the development of multimodal Gaussian splatting for dynamic scene reconstruction.

\section*{Acknowledgments}
This work was supported by the National Natural Science Foundation of China under Grant U24B20139.

{
  \small
  \bibliographystyle{ieeenat_fullname}
  \bibliography{egbib}

\begin{thebibliography}{10}

\bibitem{mildenhall2021nerf}
Ben Mildenhall, Pratul~P Srinivasan, Matthew Tancik, Jonathan~T Barron, Ravi Ramamoorthi, and Ren Ng.
\newblock Nerf: Representing scenes as neural radiance fields for view synthesis.
\newblock {\em Communications of the ACM}, 65(1):99--106, 2021.

\bibitem{kerbl20233d}
Bernhard Kerbl, Georgios Kopanas, Thomas Leimk{\"u}hler, and George Drettakis.
\newblock 3d gaussian splatting for real-time radiance field rendering.
\newblock {\em ACM Trans. Graph.}, 42(4):139--1, 2023.

\bibitem{fang2022fast}
Jiemin Fang, Taoran Yi, Xinggang Wang, Lingxi Xie, Xiaopeng Zhang, Wenyu Liu, Matthias Nie{\ss}ner, and Qi~Tian.
\newblock Fast dynamic radiance fields with time-aware neural voxels.
\newblock In {\em SIGGRAPH Asia 2022 Conference Papers}, pages 1--9, 2022.

\bibitem{yang2024real}
Zeyu Yang, Hongye Yang, Zijie Pan, and Li~Zhang.
\newblock Real-time photorealistic dynamic scene representation and rendering with 4d gaussian splatting.
\newblock In {\em Int. Conf. Learn. Represent.}, 2024.

\bibitem{lin2024gaussian}
Youtian Lin, Zuozhuo Dai, Siyu Zhu, and Yao Yao.
\newblock Gaussian-flow: 4d reconstruction with dynamic 3d gaussian particle.
\newblock In {\em IEEE Conf. Comput. Vis. Pattern Recog.}, pages 21136--21145, 2024.

\bibitem{wu20244d}
Guanjun Wu, Taoran Yi, Jiemin Fang, Lingxi Xie, Xiaopeng Zhang, Wei Wei, Wenyu Liu, Qi~Tian, and Xinggang Wang.
\newblock 4d gaussian splatting for real-time dynamic scene rendering.
\newblock In {\em IEEE Conf. Comput. Vis. Pattern Recog.}, pages 20310--20320, 2024.

\bibitem{wang2024shape}
Qianqian Wang, Vickie Ye, Hang Gao, Jake Austin, Zhengqi Li, and Angjoo Kanazawa.
\newblock Shape of motion: 4d reconstruction from a single video.
\newblock {\em arXiv preprint arXiv:2407.13764}, 2024.

\bibitem{Gallego2019EventBasedVA}
Guillermo Gallego, Tobi Delbr{\"u}ck, G.~Orchard, Chiara Bartolozzi, Brian Taba, Andrea Censi, Stefan Leutenegger, Andrew~J. Davison, J{\"o}rg Conradt, Kostas Daniilidis, and Davide Scaramuzza.
\newblock Event-based vision: A survey.
\newblock {\em IEEE Trans. Pattern Anal. Mach. Intell.}, 44:154--180, 2019.

\bibitem{achanta2012slic}
Radhakrishna Achanta, Appu Shaji, Kevin Smith, Aurelien Lucchi, Pascal Fua, and Sabine S{\"u}sstrunk.
\newblock Slic superpixels compared to state-of-the-art superpixel methods.
\newblock 34(11):2274--2282, 2012.

\bibitem{einicke1999robust}
Garry~A Einicke and Langford~B White.
\newblock Robust extended kalman filtering.
\newblock {\em IEEE transactions on signal processing}, 47(9):2596--2599, 1999.

\bibitem{snavely2006photo}
Noah Snavely, Steven~M Seitz, and Richard Szeliski.
\newblock Photo tourism: exploring photo collections in 3d.
\newblock In {\em ACM siggraph 2006 papers}, pages 835--846. 2006.

\bibitem{mur2015orb}
Raul Mur-Artal, Jose Maria~Martinez Montiel, and Juan~D Tardos.
\newblock Orb-slam: a versatile and accurate monocular slam system.
\newblock {\em IEEE transactions on robotics}, 31(5):1147--1163, 2015.

\bibitem{aliev2020neural}
Kara-Ali Aliev, Artem Sevastopolsky, Maria Kolos, Dmitry Ulyanov, and Victor Lempitsky.
\newblock Neural point-based graphics.
\newblock In {\em Eur. Conf. Comput. Vis.}, pages 696--712. Springer, 2020.

\bibitem{kopanas2021point}
Georgios Kopanas, Julien Philip, Thomas Leimk{\"u}hler, and George Drettakis.
\newblock Point-based neural rendering with per-view optimization.
\newblock In {\em Computer Graphics Forum}, volume~40, pages 29--43. Wiley Online Library, 2021.

\bibitem{ruckert2022adop}
Darius R{\"u}ckert, Linus Franke, and Marc Stamminger.
\newblock Adop: Approximate differentiable one-pixel point rendering.
\newblock {\em ACM Trans. Graph.}, 41(4):1--14, 2022.

\bibitem{guo2019relightables}
Kaiwen Guo, Peter Lincoln, Philip Davidson, Jay Busch, Xueming Yu, Matt Whalen, Geoff Harvey, Sergio Orts-Escolano, Rohit Pandey, Jason Dourgarian, et~al.
\newblock The relightables: Volumetric performance capture of humans with realistic relighting.
\newblock {\em ACM Trans. Graph.}, 38(6):1--19, 2019.

\bibitem{li2017robust}
Zhong Li, Yu~Ji, Wei Yang, Jinwei Ye, and Jingyi Yu.
\newblock Robust 3d human motion reconstruction via dynamic template construction.
\newblock In {\em International Conference on 3D Vision}, pages 496--505. IEEE, 2017.

\bibitem{broxton2020immersive}
Michael Broxton, John Flynn, Ryan Overbeck, Daniel Erickson, Peter Hedman, Matthew Duvall, Jason Dourgarian, Jay Busch, Matt Whalen, and Paul Debevec.
\newblock Immersive light field video with a layered mesh representation.
\newblock {\em ACM Trans. Graph.}, 39(4):86--1, 2020.

\bibitem{collet2015high}
Alvaro Collet, Ming Chuang, Pat Sweeney, Don Gillett, Dennis Evseev, David Calabrese, Hugues Hoppe, Adam Kirk, and Steve Sullivan.
\newblock High-quality streamable free-viewpoint video.
\newblock {\em ACM Trans. Graph.}, 34(4):1--13, 2015.

\bibitem{guo2015robust}
Kaiwen Guo, Feng Xu, Yangang Wang, Yebin Liu, and Qionghai Dai.
\newblock Robust non-rigid motion tracking and surface reconstruction using l0 regularization.
\newblock In {\em Int. Conf. Comput. Vis.}, pages 3083--3091, 2015.

\bibitem{su2020robustfusion}
Zhuo Su, Lan Xu, Zerong Zheng, Tao Yu, Yebin Liu, and Lu~Fang.
\newblock Robustfusion: Human volumetric capture with data-driven visual cues using a rgbd camera.
\newblock In {\em Eur. Conf. Comput. Vis.}, pages 246--264. Springer, 2020.

\bibitem{park2021hypernerf}
Keunhong Park, Utkarsh Sinha, Peter Hedman, Jonathan~T Barron, Sofien Bouaziz, Dan~B Goldman, Ricardo Martin-Brualla, and Steven~M Seitz.
\newblock Hypernerf: A higher-dimensional representation for topologically varying neural radiance fields.
\newblock {\em arXiv preprint arXiv:2106.13228}, 2021.

\bibitem{yang2023emernerf}
Jiawei Yang, Boris Ivanovic, Or~Litany, Xinshuo Weng, Seung~Wook Kim, Boyi Li, Tong Che, Danfei Xu, Sanja Fidler, Marco Pavone, et~al.
\newblock Emernerf: Emergent spatial-temporal scene decomposition via self-supervision.
\newblock In {\em Int. Conf. Learn. Represent.}, 2024.

\bibitem{li2024spacetime}
Zhan Li, Zhang Chen, Zhong Li, and Yi~Xu.
\newblock Spacetime gaussian feature splatting for real-time dynamic view synthesis.
\newblock In {\em IEEE Conf. Comput. Vis. Pattern Recog.}, pages 8508--8520, 2024.

\bibitem{zhu2024motiongs}
Ruijie Zhu, Yanzhe Liang, Hanzhi Chang, Jiacheng Deng, Jiahao Lu, Wenfei Yang, Tianzhu Zhang, and Yongdong Zhang.
\newblock Motiongs: Exploring explicit motion guidance for deformable 3d gaussian splatting.
\newblock {\em arXiv preprint arXiv:2410.07707}, 2024.

\bibitem{rudnev2023eventnerf}
Viktor Rudnev, Mohamed Elgharib, Christian Theobalt, and Vladislav Golyanik.
\newblock Eventnerf: Neural radiance fields from a single colour event camera.
\newblock In {\em IEEE Conf. Comput. Vis. Pattern Recog.}, pages 4992--5002, 2023.

\bibitem{qi2023e2nerf}
Yunshan Qi, Lin Zhu, Yu~Zhang, and Jia Li.
\newblock E2nerf: Event enhanced neural radiance fields from blurry images.
\newblock In {\em Int. Conf. Comput. Vis.}, pages 13254--13264, 2023.

\bibitem{ma2023deformable}
Qi~Ma, Danda~Pani Paudel, Ajad Chhatkuli, and Luc Van~Gool.
\newblock Deformable neural radiance fields using rgb and event cameras.
\newblock In {\em Int. Conf. Comput. Vis.}, pages 3590--3600, 2023.

\bibitem{xiong2024event3dgs}
Tianyi Xiong, Jiayi Wu, Botao He, Cornelia Fermuller, Yiannis Aloimonos, Heng Huang, and Christopher Metzler.
\newblock Event3dgs: Event-based 3d gaussian splatting for high-speed robot egomotion.
\newblock In {\em 8th Annual Conference on Robot Learning}, 2024.

\bibitem{deguchi2024e2gs}
Hiroyuki Deguchi, Mana Masuda, Takuya Nakabayashi, and Hideo Saito.
\newblock E2gs: Event enhanced gaussian splatting.
\newblock In {\em 2024 IEEE International Conference on Image Processing (ICIP)}, pages 1676--1682. IEEE, 2024.

\bibitem{zhang2023generalizing}
Xiang Zhang, Lei Yu, Wen Yang, Jianzhuang Liu, and Gui-Song Xia.
\newblock Generalizing event-based motion deblurring in real-world scenarios.
\newblock In {\em Int. Conf. Comput. Vis.}, pages 10734--10744, 2023.

\bibitem{liang2024clusterfomer}
James Liang, Yiming Cui, Qifan Wang, Tong Geng, Wenguan Wang, and Dongfang Liu.
\newblock Clusterfomer: clustering as a universal visual learner.
\newblock {\em Adv. Neural Inform. Process. Syst.}, 36, 2024.

\bibitem{yifan2019differentiable}
Wang Yifan, Felice Serena, Shihao Wu, Cengiz {\"O}ztireli, and Olga Sorkine-Hornung.
\newblock Differentiable surface splatting for point-based geometry processing.
\newblock {\em ACM Trans. Graph.}, 38(6):1--14, 2019.

\bibitem{lassner2021pulsar}
Christoph Lassner and Michael Zollhofer.
\newblock Pulsar: Efficient sphere-based neural rendering.
\newblock In {\em IEEE Conf. Comput. Vis. Pattern Recog.}, pages 1440--1449, 2021.

\bibitem{paszke2019pytorch}
Adam Paszke, Sam Gross, Francisco Massa, Adam Lerer, James Bradbury, Gregory Chanan, Trevor Killeen, Zeming Lin, Natalia Gimelshein, Luca Antiga, et~al.
\newblock Pytorch: An imperative style, high-performance deep learning library.
\newblock {\em Adv. Neural Inform. Process. Syst.}, 32, 2019.

\bibitem{hu2021v2e}
Yuhuang Hu, Shih-Chii Liu, and Tobi Delbruck.
\newblock v2e: From video frames to realistic dvs events.
\newblock In {\em IEEE Conf. Comput. Vis. Pattern Recog. Worksh.}, pages 1312--1321, 2021.

\bibitem{scheerlinck2019ced}
Cedric Scheerlinck, Henri Rebecq, Timo Stoffregen, Nick Barnes, Robert Mahony, and Davide Scaramuzza.
\newblock Ced: Color event camera dataset.
\newblock In {\em IEEE Conf. Comput. Vis. Pattern Recog. Worksh.}, pages 0--0, 2019.

\bibitem{kupyn2018deblurgan}
Orest Kupyn, Volodymyr Budzan, Mykola Mykhailych, Dmytro Mishkin, and Ji{\v{r}}{\'\i} Matas.
\newblock Deblurgan: Blind motion deblurring using conditional adversarial networks.
\newblock In {\em IEEE Conf. Comput. Vis. Pattern Recog.}, pages 8183--8192, 2018.

\bibitem{hore2010image}
Alain Hore and Djemel Ziou.
\newblock Image quality metrics: Psnr vs. ssim.
\newblock In {\em 2010 20th international conference on pattern recognition}, pages 2366--2369. IEEE, 2010.

\bibitem{wang2004image}
Zhou Wang, Alan~C Bovik, Hamid~R Sheikh, and Eero~P Simoncelli.
\newblock Image quality assessment: from error visibility to structural similarity.
\newblock {\em IEEE Trans. Image Process.}, 13(4):600--612, 2004.

\bibitem{zhang2018unreasonable}
Richard Zhang, Phillip Isola, Alexei~A Efros, Eli Shechtman, and Oliver Wang.
\newblock The unreasonable effectiveness of deep features as a perceptual metric.
\newblock In {\em IEEE Conf. Comput. Vis. Pattern Recog.}, pages 586--595, 2018.

\bibitem{tulyakov2021time}
Stepan Tulyakov, Daniel Gehrig, Stamatios Georgoulis, Julius Erbach, Mathias Gehrig, Yuanyou Li, and Davide Scaramuzza.
\newblock Time lens: Event-based video frame interpolation.
\newblock In {\em IEEE Conf. Comput. Vis. Pattern Recog.}, pages 16155--16164, 2021.

\end{thebibliography}
}

\end{document}